\documentclass[a4paper]{article}
\pdfoutput=1

\usepackage{IEK10} 
\usepackage{natbib}

\def\IEK10{
  Forschungszentrum Jülich GmbH,
  Institute of Energy and Climate Research,
  Energy Systems Engineering (IEK-10),
  Jülich 52425,
  Germany
}
\def\RWTH{
  RWTH Aachen University
  Aachen 52062,
  Germany
}
\def\IEKSTE{
  Forschungszentrum Jülich GmbH,
  Institute of Energy and Climate Research,
  Systems Analysis and Technology Evaluation (IEK-STE),
  Jülich 52428,
  Germany
}
\def\Colone{
    Institute for Theoretical Physics, 
    University of Cologne, 
    50937 K\"oln, 
    Germany
}
\def\DLR{
    German Aerospace Center (DLR), 
    Institute of Networked Energy Systems, 
    Oldenburg 26129, 
    Germany
}
\def\Oslo{
    Department of Computer Science, 
    OsloMet -- Oslo Metropolitan University, 
    N-0130 Oslo, 
    Norway
}
\def\JARA{
  JARA-ENERGY,
  Jülich 52425,
  Germany
}
\def\SVT{
  RWTH Aachen University,
  Process Systems Engineering (AVT.SVT),
  Aachen 52074,
  Germany
}
\def\QM{
    School of Mathematical Sciences, 
    Queen Mary University of London, 
    London E1 4NS, 
    United Kingdom
}
\def\NULS{
    Faculty of Science and Technology, 
    Norwegian University of Life Sciences, 
    1432 \AA s, 
    Norway
}

\newcommand{\mytitle}{Validation Methods for Energy Time Series Scenarios from Deep Generative Models}

\newcommand{\affil}{
  \begin{itemize}[leftmargin=3mm, itemsep=0mm]
    \item[$^a$]\IEK10
    \item[$^b$]\RWTH
    \item[$^c$]\IEKSTE
    \item[$^d$]\Colone
    \item[$^e$]\DLR
    \item[$^f$]\Oslo
    \item[$^g$]\JARA
    \item[$^h$]\SVT
    \item[$^i$]\QM
    \item[$^j$]\NULS
  \end{itemize}
}

\def\firstAuthor{Eike Cramer}
\newcommand{\myauthor}{
    \firstAuthor$^{a,b}$, 
    Leonardo Rydin Gorj\~ao$^{c,d,e,f}$, 
    Alexander Mitsos$^{g,a,h}$, 
    Benjamin Sch{\"a}fer$^{i,j}$, 
    Dirk Witthaut$^{c,d}$, 
    Manuel Dahmen$^{a,*}$ }
\newcommand{\correspondingauthor}{Manuel Dahmen, \IEK10 \newline E-mail: m.dahmen@fz-juelich.de}
\author{\myauthor}

\usepackage[
  colorlinks,
  linkcolor=blue,
  citecolor=blue,
  urlcolor=blue,
  pdftitle={\mytitle},
  pdfauthor={\firstAuthor}
]{hyperref}
\usepackage[capitalise, nameinlink]{cleveref}
\crefname{table}{Tab.}{Tab.}

\begin{document}

  \thispagestyle{firststyle}

  \begin{center}
    \begin{large}
      \textbf{\mytitle}
    \end{large} \\
    \myauthor
  \end{center}

  \vspace{0.5cm}

  \begin{footnotesize}
    \affil
  \end{footnotesize}

  \vspace{0.5cm}

  \begin{abstract}
    The design and operation of modern energy systems are heavily influenced by time-dependent and uncertain parameters, e.g., renewable electricity generation, load-demand, and electricity prices. These are typically represented by a set of discrete realizations known as scenarios.
A popular scenario generation approach uses deep generative models (DGM) that allow scenario generation without prior assumptions about the data distribution. 
However, the validation of generated scenarios is difficult, and a comprehensive discussion about appropriate validation methods is currently lacking.
To start this discussion, we provide a critical assessment of the currently used validation methods in the energy scenario generation literature. 
In particular, we assess validation methods based on probability density, auto-correlation, and power spectral density. Furthermore, we propose using the multifractal detrended fluctuation analysis (MFDFA) as an additional validation method for non-trivial features like peaks, bursts, and plateaus.
As representative examples, we train generative adversarial networks (GANs), Wasserstein GANs (WGANs), and variational autoencoders (VAEs) on two renewable power generation time series (photovoltaic and wind from Germany in 2013 to 2015) and an intra-day electricity price time series form the European Energy Exchange in 2017 to 2019.
We apply the four validation methods to both the historical and the generated data and discuss the interpretation of validation results as well as common mistakes, pitfalls, and limitations of the validation methods.
Our assessment shows that no single method sufficiently characterizes a scenario but ideally validation should include multiple methods and be interpreted carefully in the context of scenarios over short time periods. 

  \end{abstract}
    
  \vspace{0.5cm}
  
  \noindent \textbf{Keywords}:\\\textit{renewable energy; scenario generation; normalizing flows; dimensionality reduction; principal component analysis}

  \section{Introduction}\label{sec:Introduction}
The design and operation of modern energy systems are subject to uncertainties from many different sources~\citep{mitsos2018challenges}.
Renewable electricity generation from wind turbines and solar PV panels depends on the weather and fluctuates on multiple time scales~\citep{anvari2016short}. 
The integration of new consumers and different sectors increases both the complexity and the flexibility of the system on the demand side~\citep{brown2018synergies}. 
Electricity prices in liberalized markets show strong fluctuations reflecting the increasing variability of supply and demand~\citep{wolff2017short,markle2018contract}. Balancing these fluctuations is a central challenge in the transition to a future sustainable energy system. 
Hence, optimal design and operation of the energy system crucially depend on an accurate quantification and representation of the different stochastic processes~\citep{morales2013integrating,conejo2010decision,schafer2019model}.
For numerical optimization, the distributions of stochastic processes are often discretized in the form of scenarios. Each scenario then describes a possible realization over a given time period~\citep{morales2013integrating,conejo2010decision}, such that the (relatively small) collection of scenarios accurately represents the distribution~\citep{kaut2003evaluation,heitsch2003scenario}. 
In most cases, energy time series exhibit highly correlated time steps and do not follow standard distributions. 
Thus, scenario generation requires specialized methods to sample from a stochastic process. 

Recent research in scenario generation for energy time series focuses on deep generative models (DGMs), i.e., deep artificial neural networks (ANNs) that implicitly learn the distribution of a set of training data and allow for sampling without making assumptions about the underlying distribution~\citep{bond2021deep}.
Two prominent DGMs for scenario generation are $(i)$ variational autoencoders (VAEs)~\citep{kingma2013auto} and $(ii)$ generative adversarial networks (GANs)~\citep{goodfellow2014generative}. A popular modification of GANs are so-called Wasserstein GANs (WGANs), i.e., to use Wasserstein loss functions~\citep{arjovsky2017wasserstein}.
Applications of VAEs and (W)GANs include learning distributions of PV and wind power generation~\citep{zhanga2018optimized,chen2018model,jiang2018scenario,wei2019short,zhang2020typical,chen2018bayesian,jiang2019forecasting,schreiber2019generative}, concentrated solar power generation~\citep{qi2020optimal}, electric vehicle power demand~\citep{pan2019data}, and residential load~\citep{gu2019gan}. 

Both VAEs and (W)GANs use unsupervised learning algorithms, and their loss functions do not explicitly enforce the generation of realistic data. 
Thus, to ensure a reliable representation of the stochastic processes, it is crucial to assess whether the generated data shares the same characteristics as the original data.
For energy scenarios, this validation must rely on methods giving relevant insight into the essential characteristics of time series data.
Furthermore, validation methods must be applied and their results must be interpreted correctly, in particular, for time series over short time periods, to ensure comparable and reliable results. 
While there is an ongoing discussion on validation of other types of data such as generated images~\citep{salimans2016improved,borji2018pros}, we are not aware of a comprehensive discussion about the validation of DGM-generated energy time series scenarios.

To start this discussion, we critically assess the state of the art of application of validation methods and interpretation of their results in the DGM-based energy scenario generation literature. 
To this end, we review the application of the most commonly used validation methods for energy time series scenarios, namely probability density function (PDF), autocorrelation function (ACF), and power spectral density (PSD). 
For the analysis of more complex features like peaks, bursts, and plateaus, we propose using the multifractal detrended fluctuation analysis (MFDFA) as an additional validation method. 
Furthermore, we highlight shortcomings of the validation methods and explicate frequently encountered pitfalls in the interpretation of the validation results, in particular, in the context of scenarios over short time periods. 
Our assessment specifically focuses on the application of the validation methods in the context of independent and identically distributed (iid) DGM-generated scenarios. Note that other generation approaches, e.g., approaches that consider a day-ahead setting where there is additional information available, may require different validation methods.
In our numerical experiments, we train GANs, WGANs, and VAEs on scenarios of three historical energy time series, i.e., PV and wind power generation as well as intra-day price.
The validation results of the generated scenario are then compared to the historical data. 
We use the generated data to illustrate a proper application and interpretation of the validation methods and debate possible misconceptions about the information retrieved by the validation methods. 
We emphasize that our evaluation does not aim to assess whether the considered DGMs can reproduce the specific features of the data under consideration. Instead, we aim to assess whether the validation methods allow for the conclusions often made in the literature.
While our assessment focuses on scenario generation in the field of energy systems, all considered validation methods apply to any iid time series scenario set in general and therefore can be applied in other domains as well. 

Throughout this paper, we use the following notation: 
Scalars are denoted using light face (e.g., $x$) and vectors are denoted in boldface (e.g., $\mathbf{x}$). 
ANNs are denoted using boldface and capital letters (e.g., $\mathbf{G}$).
Any values generated or inferred by an ANN are marked with a tilde (e.g., $\tilde{x}$). We use the subscript $\bm{\theta}$ for distribution functions that result from an ANN transformations (e.g., $p_{\bm{\theta}}(\tilde{\mathbf{x}})$).
Stochastic processes are written as a function of time $x(t)$ and the corresponding time series is denoted using an index $x_t$. We use the term scenario to refer to one particular realization, e.g., a time series with a length of one day.

The remainder of this paper is organized as follows. 
Section~\ref{sec:dgm} reviews the fundamental theory of GANs, WGANs, and VAEs, and briefly discusses their application in generating time series scenarios. 
In Section~\ref{sec:Theory}, we review the validation methods and critically assess their application in the energy scenario generation literature. 
In Section~\ref{sec:Example}, we apply the validation methods to generated scenarios and use the results to explicate common pitfalls and explain best practices.
Finally, Section~\ref{sec:Conclusion} provides a general discussion, raises open research questions, and concludes our work. 

  \section{Deep Generative Models}\label{sec:dgm}
In this section, we present the two most widely applied DGMs, namely VAEs and GANs with their modification to WGANs.
In particular, we present the structural setup and the loss functions of the DGMs. 
For more comprehensive introductions, we refer to the original publications on GANs \citep{goodfellow2014generative}, WGANs \citep{arjovsky2017wasserstein}, and VAEs \citep{kingma2013auto}.

\subsection{Generative Adversarial Networks}
\begin{figure}
    \centering
    \includegraphics[width=0.7\linewidth]{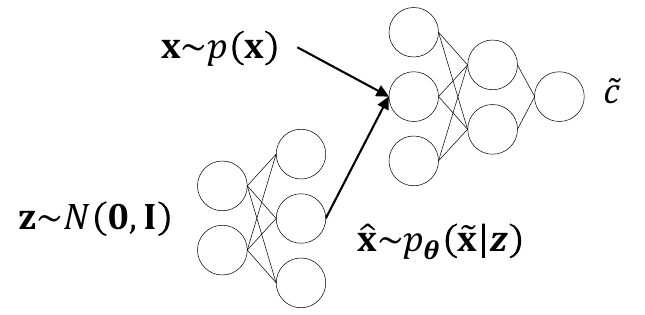}
    \caption{
    Generative adversarial network (GAN), 
    with sampling distribution $\mathbf{z}\sim N(\mathbf{0}, \mathbf{I})$, 
    true data distribution $\mathbf{x}\sim p(\mathbf{x})$,
    generated data distribution $\tilde{\mathbf{x}}\sim p_{\bm{\theta}}(\tilde{\mathbf{x}}\vert \mathbf{z})$, 
    and the classification by the discriminator $\tilde{c}_{1/0}$.}
    \label{fig:gan}
\end{figure}

GANs were proposed by \cite{goodfellow2014generative}. 
Their setup consists of two ANNs called generator $\mathbf{G}(\mathbf{z}; \bm{\bm{\theta}_G} ) $ and discriminator $\mathbf{D}(\mathbf{x};\bm{\theta}_\mathbf{D})$ which are parameterized by the parameters $\bm{\theta}_\mathbf{G}$ and $\bm{\theta}_\mathbf{D}$, respectively. 
The generator is trained to transform samples of multivariate white noise $\mathbf{z}\sim N(\mathbf{0}, \mathbf{I})$, where $\mathbf{0}$ is a vector of zeros and  $\mathbf{I}$ is the identity matrix, to data $\tilde{\mathbf{x}}$ that follows the distribution of the training data $\mathbf{x}\in X$ without learning the distribution $p_X(\mathbf{x})$ explicitly:
\begin{align}
	\mathbf{\tilde{x}} = \mathbf{G}(\mathbf{z}; \bm{\theta}_\mathbf{G})
\end{align}
The generated data then follows the conditional distribution $\tilde{\mathbf{x}}\sim p_{\bm{\theta}}(\tilde{\mathbf{x}}\vert \mathbf{z})$, where we use the subscript $\bm{\theta}$ to indicate that the distribution is based on an ANN transformation. 
The discriminator acts as an adversary to the generator. It is trained to classify between real data $\mathbf{x}$ and generated data $\mathbf{\tilde{x}}$:
\begin{align}
	\tilde{c}_{1/0} &= \mathbf{D}(\mathbf{x}; \bm{\theta}_\mathbf{D}) \label{eq:GANdiscriminator}
\end{align}
In Equation~\eqref{eq:GANdiscriminator}, the output $\tilde{c}_{1/0}$ is the classification of the data where a value of $1$ refers to real and $0$ refers to fake samples.
A sketch of the structural setup of a GAN is displayed in Figure~\ref{fig:gan}.

Generator and discriminator are trained in an iterative loop. 
First, the generator samples new data which is then fed into the discriminator. 
The generator loss function is the negative expectation of the discriminator output, i.e., it quantifies the performance of the generator `fooling' the discriminator:
\begin{align}
	\mathcal{L}_\mathbf{G} (\bm{\theta}_G) = - \mathbb{E}_Z\left[\mathbf{D}(\mathbf{G}(\mathbf{z};\bm{\theta}_\mathbf{G}); \bm{\theta}_\mathbf{D})\right] \label{eq:LossGenerator}
\end{align}
In Equation~\eqref{eq:LossGenerator}, $\mathbb{E}_Z$ is the expected value over $Z$.
In the next part of the training loop, the discriminator is trained to distinguish between real and fake samples by means of supervised training. 
Its loss function is designed to give large values for false classifications and vice versa:
\begin{align}
	\mathcal{L}_\mathbf{D} (\bm{\theta}_D)  =  - \mathbb{E}_X\left[\mathbf{D}(\mathbf{x}; \bm{\theta}_\mathbf{D})\right]  + \mathbb{E}_Z\left[\mathbf{D}(\mathbf{G}(\mathbf{z};\bm{\theta}_\mathbf{G}); \bm{\theta}_\mathbf{D})\right] \label{eq:LossDiscriminator}
\end{align}
Since the expression $\mathbb{E}_X\left[\mathbf{D}(\mathbf{x}; \bm{\theta}_\mathbf{D})\right]$ is not influenced by the generator training, Equations \eqref{eq:LossGenerator} and \eqref{eq:LossDiscriminator} can be combined to a min-max game also called adversarial training:
\begin{align}
	\underset{\bm{\theta}_\mathbf{G}}{\min} ~\underset{\bm{\theta}_\mathbf{D}}{\max} \quad\mathbb{E}_X\left[\mathbf{D}(\mathbf{x}; \bm{\theta}_\mathbf{D})\right]  - \mathbb{E}_Z\left[\mathbf{D}(\mathbf{G}(\mathbf{z};\bm{\theta}_\mathbf{G}); \bm{\theta}_\mathbf{D})\right]
\end{align}
The min-max game continues until the generator-discriminator pair reaches a Nash equilibrium, i.e., neither generator nor discriminator can improve without the other worsening.
When the two networks are in Nash equilibrium, the generator is thought to reproduce the true distribution of the training data.

\subsection{Wasserstein-GAN}
In practice, standard GANs often converge to unstable Nash equilibria and show signs of mode dropping, i.e., a lack of diversity in the generated data \citep{metz2017unrolled,arjovsky2017principled}. 
Furthermore, \cite{arjovsky2017principled} proved that the Nash equilibrium of the traditional training approach by \cite{goodfellow2014generative} does not represent the real target distribution.

To mitigate these issues, a popular modification to the GAN is to use the Wasserstein distance as a loss function \citep{arjovsky2017wasserstein}. It describes the least amount of work to transfer one distribution into another. In practice, the Wasserstein distance itself is never explicitly enforced. Instead, the WGAN relaxes the classification approach of the discriminator to a continuous output of a so-called critic: 
\begin{align}
    \tilde{c}=\mathbf{C}(\mathbf{x};\bm{\theta}_\mathbf{C}) \label{Eq:Critic}
\end{align}
In Equation \eqref{Eq:Critic}, $\mathbf{C}(\mathbf{x};\bm{\theta}_\mathbf{C})$ is the critic with parameters $\bm{\theta}_\mathbf{C}$ and $\tilde{c}$ is a continuous confidence $\tilde{c}\in \mathbb{R}$ expressed by the critic on whether the data is real or fake. 

Employing a critic with a real-valued output instead of a discriminator with a $\{0,1\}$ classification avoids potential issues of vanishing gradients due to the sigmoid output activation of the discriminator and thereby leads to a more stable convergence \citep{arjovsky2017wasserstein}.
To enforce the Wasserstein distance, the critic has to be restricted to be Lipschitz continuous. \cite{arjovsky2017wasserstein} propose enforcing the Lipschitz constraint through weight clipping. However, other approaches using weight penalties have been proposed as well \citep{gulrajani2017improved}. For more information about enforcing the Lipschitz constraint, the reader is referred to the original publication \citep{arjovsky2017wasserstein}.

\subsection{Variational Autoencoders}
\begin{figure}
    \centering
    \includegraphics[width=0.7\linewidth]{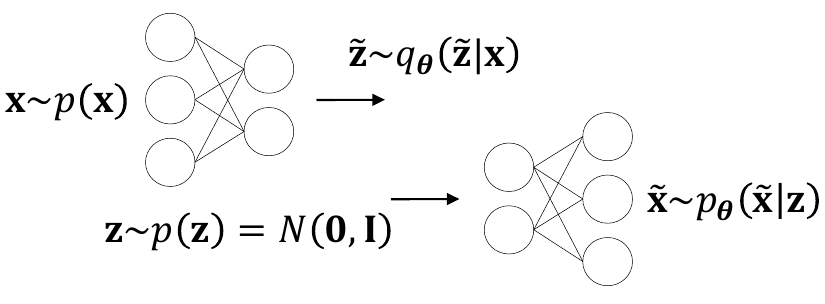}
    \caption{
        Variational autoencoder (VAE), with 
        true data distribution $\mathbf{x}\sim p(\mathbf{x})$, 
        generated data distribution $\tilde{\mathbf{x}}\sim p_{\bm{\theta}}(\tilde{\mathbf{x}}\vert\mathbf{z})$, 
        sampling/latent distribution $p(\mathbf{z})\sim N(\mathbf{0}, \mathbf{I})$, 
        and inferred latent space distribution $q_{\bm{\theta}}(\tilde{\mathbf{z}}\vert \mathbf{x})$.
        }
        \label{fig:vae}
\end{figure}

VAEs were first proposed by \cite{kingma2013auto}.
The basic idea of VAEs is to extend the concept of autoencoders to a DGM. 
Autoencoders, in general, are (non-)linear transformations that compress the data to a lower-dimensional latent space and then rebuilt the data from the encoding. The parts of the autoencoder for encoding and decoding are called encoder and decoder, respectively, and are often built using ANNs:
\begin{align}
    \tilde{\mathbf{z}} &= \mathbf{E} (\mathbf{x}; \bm{\theta}_\mathbf{E})\\
    \tilde{\mathbf{x}} &= \mathbf{D} (\tilde{\mathbf{z}}; \bm{\theta}_\mathbf{D})
\end{align}
Here, $\mathbf{x}\in X$ is the data, $\tilde{\mathbf{x}} \in \tilde{X}$ is the reconstructed data, $\tilde{\mathbf{z}} \in \tilde{Z}$ is the inferred latent space variable, and $\mathbf{E} (\mathbf{x}; \bm{\theta}_\mathbf{E})$ and $\mathbf{D}(\mathbf{z}; \bm{\theta}_\mathbf{D})$ are encoder and decoder networks with parameters $\bm{\theta}_\mathbf{E}$ and $\bm{\theta}_\mathbf{D}$, respectively.
VAEs work similarly to regular autoencoders. However, their latent space is regularized to a multivariate Gaussian. 
Thereby, VAEs can generate new data by sampling from the known latent space distribution and transforming the samples with the decoder, similar to the GAN generator.
A sketch of the structural setup of VAEs is shown in Figure~\ref{fig:vae}.

Ideally, VAEs would be trained via likelihood maximization. However, the nonlinear encoder and decoder function make the likelihood function intractable. 
Thus, a lower bound on the likelihood, also known as evidence lower-bound (ELBO), is used as the loss function:
\begin{equation}
    \begin{aligned}
        \text{ELBO}(\bm{\theta}_{\mathbf{E}},\bm{\theta}_{\mathbf{D}})~=~& 
        \mathbb{E}_{\tilde{Z}} 
        \left[ 
        \log~p_{\bm{\theta}}(\tilde{\mathbf{x}}\vert \tilde{\mathbf{z}})
        \right] \\
        & - D_{KL} \left[
        q_{\bm{\theta}}(\tilde{\mathbf{z}}\vert \mathbf{x})
        \vert\vert 
        p(\mathbf{z}) 
        \right] \label{Eq:ELBO}
    \end{aligned}
\end{equation}
Here, $p_{\bm{\theta}}(\tilde{\mathbf{x}}\vert \tilde{\mathbf{z}})$ is the distribution of the reconstructed data $\tilde{\mathbf{x}} = \mathbf{D} (\tilde{\mathbf{z}}; \bm{\theta}_\mathbf{D})$, i.e., the decoder output given a prior encoding $\tilde{\mathbf{z}} = \mathbf{E}(\mathbf{x}; \bm{\theta}_\mathbf{E})$, $D_{KL}\left[\cdot\vert\vert\cdot\right]$ is the Kullback-Leibler divergence, a measure of similarity between distributions, $q_{\bm{\theta}}(\tilde{\mathbf{z}}\vert \mathbf{x})$ is the latent space distribution given the data $\mathbf{x}$, and $p(\mathbf{z})=N(\mathbf{0},\mathbf{I})$ is a multivariate standard Gaussian. 
The first term of the ELBO in Equation \eqref{Eq:ELBO} maximizes the likelihood of the generated data. The second part with the Kullback-Leibler divergence enforces the standard Gaussian requirement for the latent space by minimizing the divergence between the inferred latent space distribution $q_{\bm{\theta}}(\tilde{\mathbf{z}}\vert \mathbf{x})$ and the target latent distribution $p(\mathbf{z})=N(\mathbf{0},\mathbf{I})$. 

Since the initial publication by \cite{kingma2013auto}, extensions to the VAE concept were proposed that mostly concern the latent space distribution and propose Gaussian mixture models \citep{guo2020variational} and normalizing flows \citep{rezende2016variational}.

\subsection{DGM-based scenario generation}\label{sec:dgmApplications}
There are multiple works that propose different versions of the different DGMs reviewed above for energy scenario generation. 
Applications of GANs and WGANs include wind power generation \citep{jiang2018scenario,zhang2020typical}, PV power generation \citep{chen2018model, chen2018bayesian, jiang2019forecasting, wei2019short, schreiber2019generative}, and residential loads \citep{gu2019gan}. 
VAEs were applied to learn the distributions of PV and wind power generation \citep{zhanga2018optimized}, concentrated solar power \citep{qi2020optimal}, and electric vehicle power demand \citep{pan2019data}.

In all works on DGM-based scenario generation, the historical time series are cut into segments of equal length of typically one day. These segments, i.e., the historical scenarios, are then viewed as multidimensional iid data points, and the DGMs are trained to sample from the distribution of these data points.

  \section{Theory of Validation Methods}\label{sec:Theory}
In this section, we briefly review PDF, ACF, PSD, and MFDFA and explain their application as validation methods for DGM-generated scenarios. Furthermore, we critically assess the application of PDF, ACF, and PSD-based validation of DGM-generated energy scenarios in the literature.

\subsection{Probability Density Function}\label{sec:TheoryPDF}
The PDF describes the likelihood of a particular realization of a continuous random variable. For arbitrary data sets, there often exists no analytical distribution model and the PDF has to be estimated, e.g., by the non-parametric kernel density estimation (KDE)~\citep{parzen1962estimation,davis2011remarks}. 
For a finite set of univariate samples, KDE assigns a kernel function $K(\cdot)$, i.e., a non-negative function that integrates to 1, to the data. The density estimate is then given by the average over all $N$ samples:
\begin{equation}
    \text{PDF} (x) = \frac{1}{hN}\sum_{i=1}^{N} K\left(\frac{x-x_i}{h}\right) \label{eq:KDE}
\end{equation}
In Equation~\eqref{eq:KDE}, $x_i$ are the samples,  $x$ is the uncertain variable, and $h$ is a smoothing parameter called bandwidth.
The most common kernel function is the Gaussian kernel~\citep{davis2011remarks}:
\begin{equation}
    \text{PDF} (x) = \frac{1}{\sqrt{2\pi}hN}\sum_{i=1}^{N} \exp{\left(-\frac{1}{2}\left(\frac{x-x_i}{h}\right)^2\right)} 
\end{equation}
Equation~\eqref{eq:KDE} gives an explicit expression, which makes computing the PDF very efficient. 

For DGM-generated scenarios, KDE can estimate either the overall PDF or the marginal PDF of the scenarios. 
For the PDF of all time steps, the time steps $x_t$ become the samples $x_i$ in Equation~\eqref{eq:KDE}. 
The marginals are the sum over all the dimensions of a multivariate distribution, i.e., the sum over all time steps of a given scenario. Hence, the marginal PDF divided by the number of time steps describes the distribution of the daily mean.

In the literature on energy scenario generation, most authors evaluate the PDF on a linear scale~\citep{gu2019gan, jiang2018scenario, jiang2019forecasting, schreiber2019generative, wang2018conditional, zhang2020typical, wei2019short,qi2020optimal, pan2019data, zhanga2018optimized} or the integral over the PDF, i.e., the cumulative distribution function (CDF) \citep{chen2018model,jiang2019forecasting}.
However, the linear scaled PDF and the CDF can only show differences between historical and generated scenarios for values of high likelihood. Thus, they hide potential mismatches in low-density regions, e.g., for electricity price distributions with heavy tails~\citep{uniejewski2017variance}. 
If the distributions are known to have heavy tails, the PDF should therefore also be investigated on a logarithmic scale to highlight the low-density regions. 

Analyzing the marginal distributions reveals additional insight into the generated scenarios that cannot be extracted from the regular PDF, e.g., the full width of seasonal changes in PV or the distribution of high and low-price days. Still, most articles do not assess the marginal distribution, with the notable exception of the work in~\citep{ge2020modeling}.

\subsection{Autocorrelation Function}\label{sec:TheoryACF}
The ACF describes the internal correlation of a stochastic process $x(t)$ with a delayed version of itself $x(t-\tau)$. For discrete time series $x_t$, the ACF is a particular case of covariance describing the correlation between different time steps~\citep{stoica2005spectral}.
The auto-covariance $K_{xx}$ of a stationary time series is given by:
\begin{equation}
    K_{xx}(\tau) = \mathbb{E}\left[(x_t -\mu_x)(x_{t+\tau}-\mu_x)\right] \label{eq:Auto_Covariance}
\end{equation}
In Equation~\eqref{eq:Auto_Covariance}, $\mu_x$ is the mean of the time series $x_t$ and $\tau$ the time-lag between the time steps. 
For the ACF, we consider the normalized auto-covariance, also known as the Pearson correlation
\begin{equation}
    R_{xx}(\tau) = \frac{\mathbb{E}\left[(x_t -\mu_x)(x_{t+\tau}-\mu_x)\right]}{\sigma_t^2}\quad \in [-1,1], \label{eq:Pearsoncorrelation}
\end{equation}
where $\sigma_t^2$ is the variance of the time series. 
The ACF has at least one maximum at $R_{xx}(\tau=0)=1$, where the expected value coalesces to the variance.
Note that different normalizations for the auto-correlation function exist. However, in this work we refer to the Pearson correlation, Equation~\eqref{eq:Pearsoncorrelation}, as the ACF.

For the validation of iid DGM-generated scenarios, it is important to note that a concatenation of multiple scenarios would result in potentially unrealistic sequences. Therefore, ACF can only evaluate single scenarios, and an analysis of the entire data set is not possible. The limitation to short single scenarios also means that the ACF can be computed very efficiently. 
The literature approach most often used randomly selects a set of 3 to 4 historical scenarios and then searches the generated scenario set for the scenarios with the most similar ACF using the Euclidean distance as a metric, i.e., the ACF trajectory is viewed as a multidimensional point and the generated scenario with the ACF closest to the ACF of the historical scenario is selected as representative of the entire scenario set. 
Most authors then go on to interpret a good match of the ACFs to indicate a correct match of the temporal behavior by the DGM-generated scenarios~\citep{chen2018model,jiang2018scenario,wei2019short,jiang2019forecasting,zhang2020typical,gu2019gan,ge2020modeling}. 

Opposed to this general understanding, we advise that the ACF results should be interpreted with caution.
First, the typical comparison of ACFs of 3 to 4 historical scenarios to the best matching generated scenarios does not give any general information about the full generated scenario set since most of the scenarios are excluded from the evaluation. On the contrary, potential outliers that may show very different ACFs are systematically excluded by the best-match comparison. 
Second, a match of ACFs does not prove that two time series are the same or even similar, e.g., two time series can stem from different stochastic processes and still show the same ACF (c.f.~Section~\ref{sec:autocorrelation}). 
In conclusion, we find that the ACF analysis approach widely used in the literature can result in potentially spurious implications about the generated data and should therefore be treated with caution.

\subsection{Power Spectral Density}\label{sec:TheoryPSD}
Stochastic processes often exhibit periodic behavior, which can be analyzed in the frequency domain using the PSD. 
The PSD describes the distribution of the fluctuational power over a range of frequencies~\citep{stoica2005spectral}. 
Here, the term power refers to the square of the signal and is distinct from the unit of the stochastic process, e.g., renewable power generation. 
The general form of the PSD $S_{xx}$, based on the Fourier transform $\hat{x}(f)$, is given by:
\begin{equation}
    S_{xx}(f) = \lim_{T\to \infty} \frac 1 {T}  \vert \hat{x}_{T}(f)\vert^2
\end{equation}
Here, $f$ is the frequency, $\vert\cdot\vert$ denotes the absolute value, and $\hat{x}_{T}(f)$ is the Fourier transform over a variable interval $T$ selected through an indicator function $I_T$, $\hat{x}_T(t)=\hat{x}(t)\cdot I_T$. The Fourier transform is given by:
\begin{equation}
    \hat{x}(f) = \int\displaylimits_{-\infty}^{\infty} x(t)\exp(-2\pi jft)\mathrm{d}t \label{eq:FourierTransform}
\end{equation}
In Equation~\eqref{eq:FourierTransform}, $j$ is the imaginary unit.
For practical applications, the PSD is computed using the fast Fourier transformation~\citep{heideman1985gauss}, which allows for very efficient computation. For a more detailed introduction, we refer to the literature~\citep{stoica2005spectral}.
The PSD of real-world time series is often noisy and difficult to read. Therefore, it is common to employ smoothing functions such as Welch's method~\citep{welch1967use}, which computes the PSD over a set of overlapping segments instead of treating the time series as a whole. The average over these segments then gives a smooth estimate of the PSD. 

The PSD is designed for consecutive time series. For the application of PSD to DGM-generated scenarios, all scenarios are concatenated to form a single time series.
Hence, the periods relevant to the validation are between the Nyquist frequency, i.e., the shortest period supported by the PSD~\citep{stoica2005spectral}, and the scenario length. 
The PSD of any period longer than the scenario length describes the concatenation of scenarios, i.e., its use is inconsequential to the aim of validation. 
However, most authors present the PSD of longer periods~\citep{chen2018model, wei2019short,qi2020optimal}. 
For instance, the authors in \cite{chen2018model} generate 24\,h scenarios in 5\,min resolution, but present the PSD over periods between 6\,d and 1\,h, which neglects short periods reflecting the short-term behavior.
Therefore, we emphasize that the presented periods must reflect the scenario length and not truncate the short-term behavior.

\subsection{Multifractal Detrended Fluctuation Analysis}
ACF and PSD form the basis of scenario validation in many contributions. However, the information they extract is limited, i.e., ACF can only quantify the correlations between time steps within a time series and PSD uncovers power-law decays of the spectrum. However, the time series data might include nontrivial features like peaks, bursts, and plateaus that are not represented in the ACF and PSD analysis, but still should be validated in order to gain confidence in the generated scenarios. To retrieve these features, we propose to use multifractal analysis.

Multifractal analysis aims to uncover the fractal composition of a stochastic process $x(t)$, i.e., the change in the fluctuation behavior relative to the sampling rate and the considered interval~\citep{Salat2017Multifractal}. The term fractal refers to a process where the whole behaves similarly to parts of itself and is therefore also known as a self-similar process.
A stochastic process with constant fluctuation behavior, e.g., Brownian motion, consists of a single fractal and is therefore called monofractal~\citep{Salat2017Multifractal}. 
If the fluctuation behavior changes with the sampling rate and/or the considered interval, the process is a composite of multiple fractals and hence called multifractal. 
For the fractal analysis, the stochastic process has to be dissected into its different scales. 
There are multiple algorithms to analyze the fractality of time series, e.g., wavelet analysis~\citep{Muzy1991wavelets, Bacry1993wavelets, Muzy1994wavelets} and MFDFA~\citep{Kantelhardt2001, Kantelhardt2002, Zhou2010, Zhang2019}. We refer to \cite{Salat2017Multifractal} for a review of these methods.
For our discussion, we focus on MFDFA, due to its straightforward application to discrete time series, and utilize a modification to moving windows to analyze short time series~\citep{Zhou2010, Zhang2019}. In the following, we give a brief introduction to the MFDFA algorithm. For a more detailed explanation, we refer to the original MFDFA paper by \cite{Kantelhardt2002}.

\begin{figure}
    \centering
    \includegraphics[width=0.7\linewidth]{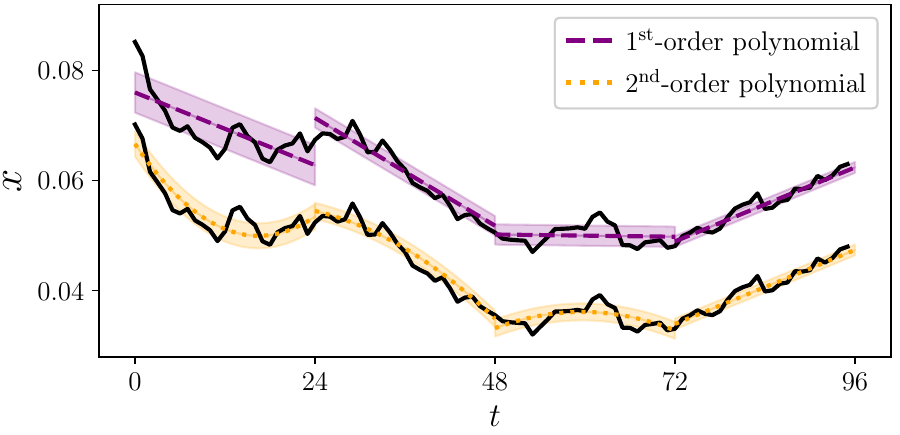}
    \caption{Wind capacity factor time series [-] with four non-overlaping windows of size $s=24$ with 1\textsuperscript{st}-order polynomial detrending (top) and 2\textsuperscript{nd}-order polynomial detrending (bottom). The standard deviation of the detrended data is shown in the shaded areas, as given by Equation~\eqref{eq:F2_poly_fits}.}
    \label{fig:mfdfa}
\end{figure}

In a first step, the MFDFA algorithm computes the centered integral $Y_i$ over the finite time series $x_k$, i.e., the cumulative sum subtracted by the mean $\mu_x$, for all time steps $i$: 
\begin{equation}\label{eq:CenteredIntegral}
    Y_i = \sum_{k=1}^i \left( x_k - \mu_x \right),\quad \forall~i=1,2, \dots, N
\end{equation}
Next, the centered integral is split into $v=1,2, \dots, N_s$ segments with a length of $s$ steps and a polynomial $y_{v}$ of order $m$ is fitted to each segment. Figure~\ref{fig:mfdfa} shows an example with first and second-order polynomials. The segmentation and polynomial fitting is repeated for a spectrum of segment lengths $s\in\{ m+2, \dots, S\}$. Hence, the number of segments $N_s$ alters with the segment length. In this work, $S$ is selected to be the number of steps in a scenario. 
To quantify the variability of the detrended process, MFDFA then computes the variance function $F^2(v,s)$ of the data around the polynomial fit for a given segment $v$ of length $s$:
\begin{equation}\label{eq:F2_poly_fits}
    F^2(v,s) = \frac{1}{s} \sum_{i=1}^{s} [Y_{(v-1)s + i} - y_{v,i}]^2, \quad \forall~v=1,2, \dots, N_s
\end{equation} 
In Equation~\eqref{eq:F2_poly_fits}, $y_{v,i}$ denotes the polynomial value at the $i$-th point of the segment. The variance function is indicated as the shaded areas around the polynomials in Figure~\ref{fig:mfdfa}.
Lastly, the MFDFA computes the $q$-th order fluctuation function $F_q(s)$ of the variance function $F^2(v,s)$ over the $N_s$ windows $v$:
\begin{equation}\label{eq:F_q2}
  F_q(s) = \left\{\frac{1}{N_s} \sum_{v=1}^{N_s} [F^2(v,s)]^{q/2}\right\}^{1/q}
\end{equation} 

MFDFA aims to retrieve the fluctuation power by determining the power scaling of the fluctuation function
\begin{equation}
    F_q(s)\sim s^{h(q)},
\end{equation}
where the exponent $h(q)$ is also known as the generalized Hurst coefficient \citep{Kantelhardt2002}. To investigate $h(q)$, the fluctuation function $F_q(s)$ is displayed over the segment length $s$ in double logarithmic scaling, where $h(q)$ is the slope of $F_q(s)$.
Parallel curves for different $q$-powers, i.e., constant Hurst coefficients $h(q)$, indicate a monofractal time series. A change in the slope for different $q$-powers indicates multifractal behavior. In general, low $q$-powers describe the mean fluctuation, and high $q$-powers extract bursts and peaks. Negative $q$-powers analyze plateaus and extended periods of low fluctuations.

To our knowledge, there is no instance of fractal analysis in the energy scenario generation literature. Therefore, we propose the following approach for MFDFA of DGM-generated scenarios:
Similar to PSD, MFDFA is designed to analyze consecutive time series. Therefore, the generated scenarios are concatenated for the evaluation. 
Then, we compute the fluctuation function using linear detrending with three positive and three negative $q$-powers.
The original MFDFA~\citep{Kantelhardt2002} computes the variance over non-overlapping segments. For the analysis of shorter time series like the DGM-generated scenarios, we adopt the sliding window approach~\citep{Zhou2010,Zhang2019}. Here, the window of length $s$ progresses step-wise, rendering a much higher number of segments $N_s$ for the analysis. 
Due to the concatenation, segments overlapping multiple scenarios are likely to connect unrealistic sequences and yield spurious results. Therefore, segment lengths exceeding half the scenario length have to be evaluated cautiously. 

As shown in Equation~\eqref{eq:F2_poly_fits}, computing the MFDFA requires fitting polynomials, which makes the computation more expensive than PDF, ACF, and PSD. In particular, using the sliding window approach leads to a high number of fitting tasks.

\subsection{Commonalities and differences}
Finally, we summarize the commonalities and differences between the four considered methods in Table~\ref{tab:CommonalitiesAndDifferences}. For instance, ACF and PSD check very similar categories with the only major difference that PSD analyzes the entire dataset. 
Importantly, only MFDFA and the PDF on logarithmic scaling analyze rare events. As neither of the two is typically used, this aspect is currently missing from the DGM-based scenario generation literature.
Except for MFDFA, all methods are computationally very efficient and can be computed within milliseconds on a personal computer. 
MFDFA computation takes up to a minute for the considered datasets and the extended computational time is mostly due to the sliding window approach.
Finally, the last line of Table~\ref{tab:CommonalitiesAndDifferences} highlights that all validation methods discussed in this paper require visual interpretations of results. These visual interpretations are potentially subjective and we hypothesize that the lack of quantitative measures contributes to the often spurious conclusions often seen in the literature. 
\begin{table}
    \caption{
        Summary of commonalities and differences of PDF, ACF, PSD, and MFDFA. 
         }
    \label{tab:CommonalitiesAndDifferences}
        \centering
        \begin{tabular}{l|cccc}
                                  & PDF             & ACF       & PSD       & MFDFA     \\ \hline
            Probability distribution & \ding{51}       & \ding{55} & \ding{55} & \ding{55} \\
            Temporal dependencies & \ding{55}       & \ding{51} & \ding{51} & \ding{51} \\
            Full data set         & \ding{51}       & \ding{55} & \ding{51} & \ding{51} \\
            Rare events           & \ding{51}$^{1}$ & \ding{55} & \ding{55} & \ding{51} \\
            Nontrivial features   & \ding{55}       & \ding{55} & \ding{55} & \ding{51} \\
            Easy interpretation   & \ding{51}       & \ding{51} & \ding{51} & \ding{55} \\
            Require concatenation & \ding{55}       & \ding{55} & \ding{51} & \ding{51} \\
            Computationally efficient & \ding{51}   & \ding{51} & \ding{51} & \ding{55} \\
            Quantitative measure  & \ding{55}   & \ding{55} & \ding{55} & \ding{55} \\
        \end{tabular}
    \\
    $^{1}$ Only on logarithmic scaling.
\end{table}
  \section{Numerical studies}\label{sec:Example}
In this section, we apply the previously discussed validation methods to DGM-generated scenarios of PV and wind capacity factors and intraday prices.
To provide a variety of different scenario sets, we utilize GAN-, WGAN-, and VAE-generated scenarios in comparison to historical data. Nevertheless, our analysis is focused solely on the assessment of the validation methods and refrains from any implications about either of the three considered DGM.  
We start by describing the considered data sets and the model structures used for the DGMs. 
Next, we apply step-by-step PDF, ACF, PSD, and MFDFA to the historical and the DGM-generated time series scenarios. 

\subsection{DGM Training}
To generate scenarios for our evaluation, we train GANs, WGANs, and VAEs on three different historical energy time series data sets. 
The data sets are: 
\begin{itemize}
    \item Daily total PV power generation in Germany from 2013 to 2015 \citep{DataSource},
    \item Daily total wind power generation in Germany from 2013 to 2015 \citep{DataSource},
    \item Intra-day price data from 2017 to 2019 from the European Energy Exchange AG (EEX), based in Leipzig, Germany, processed by the Fraunhofer Institute for Solar Energy Systems~\citep{DataSourcePrice}.
\end{itemize}
As described in Section~\ref{sec:dgmApplications}, the historical time series is cut into daily segments. Each segment is then viewed as a vector-valued iid sample of a multivariate distribution, and each time step is represented as a dimension \citep{chen2018model}. 
The time series considered here are recorded in 15\,min intervals, rendering scenarios comprised of 96\,time-steps. The collection of all scenarios then is a set of 96\,dimensional iid samples.

None of the three historical time series contains any missing or faulty time steps. The intra-day price time series contains some extreme peaks. These peaks are important parts of the process and should not be considered outliers. 
PV and wind scenarios are scaled by the total installed capacity at each time step, resulting in the so-called capacity factor.
All data sets are scaled to $[-1,1]$ for GAN and WGAN and to $[0,1]$ for the VAE to fit the respective `tanh' and `sigmoid' output activation functions.
All DGMs are implemented using the open-source, Python-based machine learning library Tensorflow, version 2.4.0~\citep{tensorflow2015}. 
Details of the respective DGM structures are listed in Table~\ref{tab:DGMs}. 
Note that our work places a special focus on the correct application and interpretation of the validation methods as opposed to focusing on generating particularly good results in the evaluation, and the design of our DGMs is not optimized via rigorous hyperparameter optimization.
\begin{table*}[ht]
\caption{DGMs: ANN layer setup for the WGAN, GAN, and VAE. 1D Conv are convolutional layers \citep{goodfellow2016deep} and 1D ConvT are transposed convolutional (deconvolutional) layers \citep{radford2016unsupervised}. The attributes for different layers are: 
Fully connected: (Number of neurons),
reshape: (output dim 1, output dim 2, \dots),
1D Convolution: (number of filters, filter size, strides, padding), and
1D Convolution Transpose: (number of filters, filter size, strides, padding).
}

\label{tab:DGMs}
\resizebox{\textwidth}{!}{
\begin{tabular}{lll|lll|lll}
\hline
\multicolumn{3}{c|}{WGAN}  & \multicolumn{3}{c|}{GAN}  & \multicolumn{3}{c}{VAE}  \\ \hline \hline
\multicolumn{3}{c|}{generator}  & \multicolumn{3}{c|}{generator}  &  \multicolumn{3}{c}{decoder}  \\ \hline
\textbf{Layer}   & \textbf{attributes}  & \textbf{activation}    & \textbf{Layer}  & \textbf{attributes}  & \textbf{activation}    & \textbf{Layer}            & \textbf{attributes}   & \textbf{activation} \\
Fully connected  & (1152)        & ReLU          & Fully connected   & (1152)        & ReLU          & Fully connected  & (1152)       & ReLU   \\
reshape          & (96, 12)       & ReLU          & reshape           & (96, 12)       & ReLU          & reshape          & (96, 12)        & ReLU   \\
1D ConvT         & (12, 3, 1, 1) & ReLU          & 1D ConvT          & (12, 3, 1, 1) & ReLU          & 1D ConvT         & (12, 3, 1, 1)        & ReLU   \\
1D ConvT         & (1, 3 ,1, 1)  & None          & 1D ConvT          & (1, 3, 1, 1)  & None          & 1D ConvT         & (1, 3, 1, 1)         & sigmoid  \\\hline

\multicolumn{3}{c|}{critic}  & \multicolumn{3}{c|}{discriminator}  &  \multicolumn{3}{c}{encoder}  \\ \hline
\textbf{Layer}  & \textbf{attributes}  & \textbf{activation} & \textbf{Layer}            & \textbf{attributes}  & \textbf{activation} & \textbf{Layer}            & \textbf{attributes}   & \textbf{activation} \\
1D Conv         & (12, 3, 1, 1) & LeakyReLU  & 1D Conv          & (12, 3, 1, 1) & ReLU       & 1D Conv          & (16, 5, 1, 0)        & ReLU   \\
1D Conv         & (4, 3, 1, 1)  & LeakyReLU  & 1D Conv          & (4, 3, 1, 1)  & ReLU       & Flatten          & (-)        & None   \\
Flatten         & (-)           & None       & Flatten          & (-)           & None       & Fully connected  & (80)         & ReLU   \\
Fully connected & (1)           & None       & Fully connected  & (1)           & tanh       & Fully connected  & (20+20)      & None \\\hline
\end{tabular}}
\end{table*}

\begin{figure}
    \centering
    \includegraphics[width=0.7\linewidth]{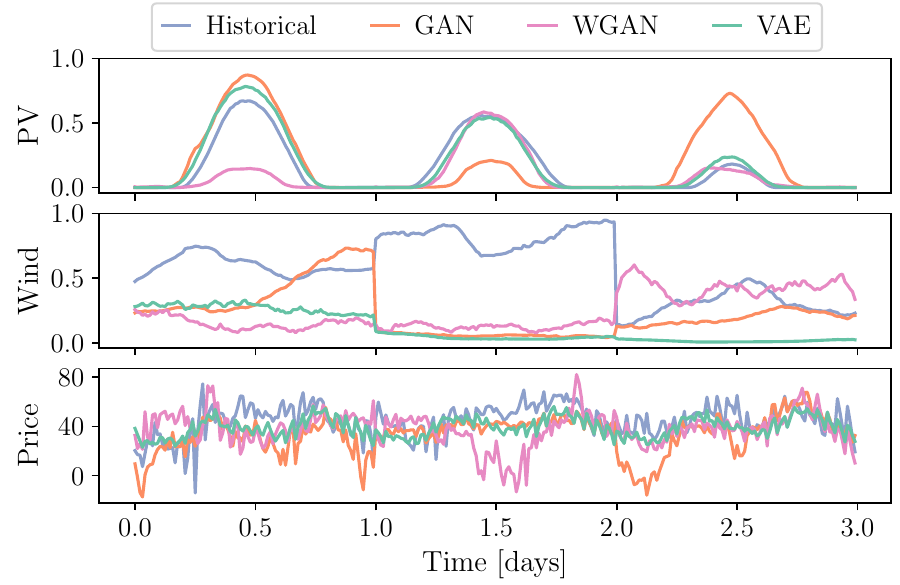}
    \caption{Examples of PV capacity factor [-], wind capacity factor [-] from Germany in 2013 to 2015, and intra-day price [EUR/kWh] time series in the European Energy Exchange in 2017 to 2019 (labels shortened in figure). Time series are concatenations of three randomly selected scenarios from historical~\citep{DataSource, DataSourcePrice} and GAN, WGAN, and VAE-generated scenario sets.}
    \label{fig:DataExample}
\end{figure}
We train GAN and VAE using the Adam optimizer~\citep{kingma2017adam} and WGAN using {RMSprop}~\citep{Tieleman2012} as recommended in the WGAN paper~\citep{arjovsky2017wasserstein}.
All networks are trained for $2000$ epochs with a learning rate of $10^{-5}$.
As the loss functions of neither (W)GAN nor VAE provide stopping criteria for the training, we take to following approach to selecting the trained models:
After every epoch we save the models, draw $1000$ samples, and evaluate their PDF and PSD, see Sections~\ref{sec:TheoryPDF} and~\ref{sec:TheoryPSD}. After the $2000$ epochs, we select the model from the epoch where the two metrics best match the historical data as judged through visual inspection. 
In Figure~\ref{fig:DataExample}, we show some exemplary data of the three different historical time series~\citep{DataSource, DataSourcePrice} and the generated scenarios.
The time series are displayed over three days and are concatenations of randomly selected scenarios each.

Next, we apply the validation methods discussed in Section~\ref{sec:Theory} to the generated scenarios.

\subsection{Probability density function}\label{sec:kerneldensity}
\begin{figure*}[t]
    \centering
    \includegraphics[width=1\textwidth]{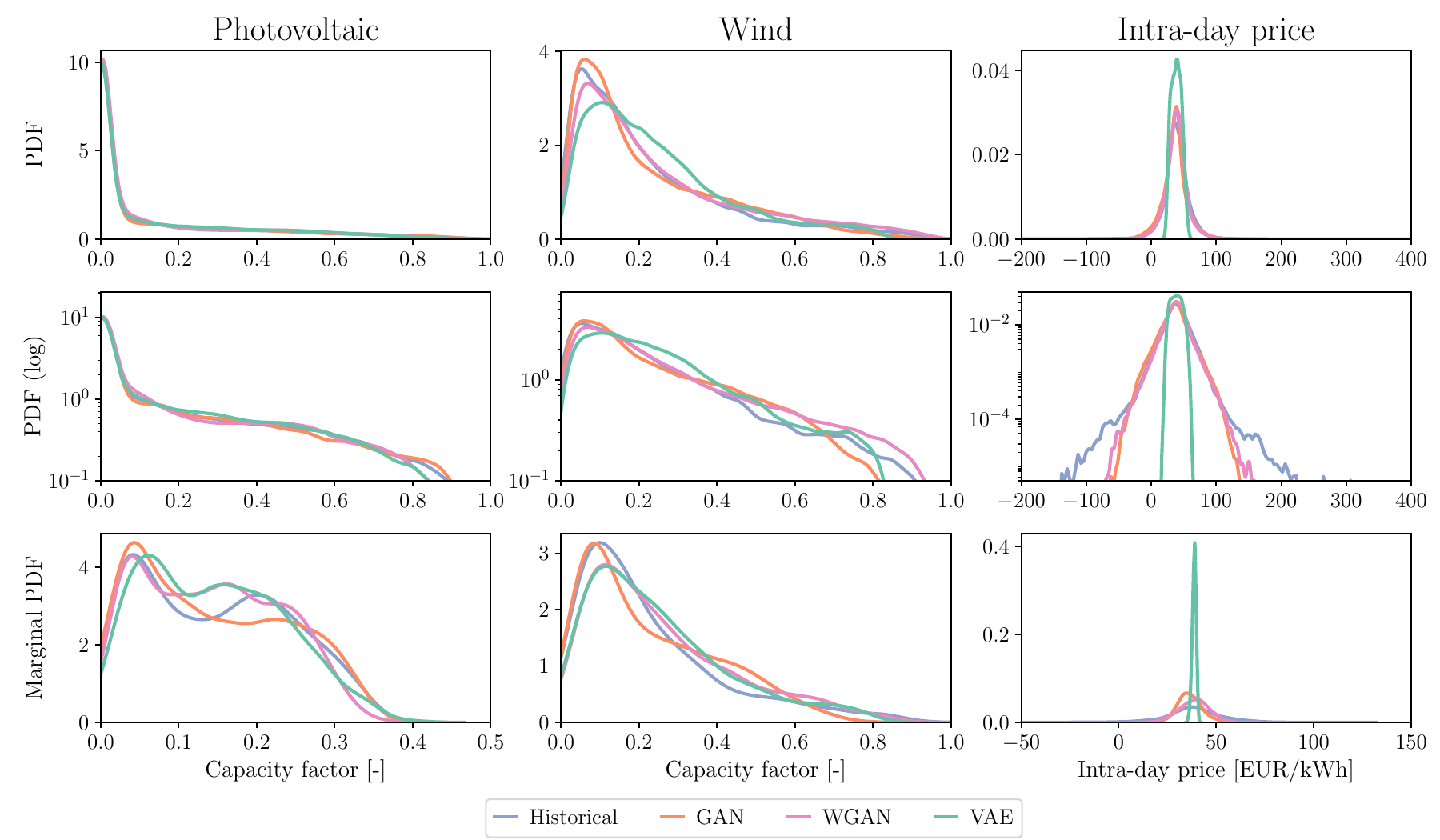}
    \caption{Kernel density estimates of the PDF of PV and wind capacity factor, and  intra-day price scenarios. Historical datasets~\citep{DataSource,DataSourcePrice} are compared to and GAN-, WGAN-, and VAE-generated scenarios.}
    \label{fig:PDF_grid}
\end{figure*}
We apply KDE with Gaussian kernels to estimate the PDF of the historical time series and the scenarios generated using GAN, WGAN, and VAE.
For completeness, we show the PDFs on linear and logarithmic scales and examine the scaled marginal distributions, i.e., the daily mean of capacity factor and intra-day price.
All PDF calculations and plots are generated using the Python-based library Seaborn, version 0.11.1~\citep{waskom2020seaborn}, and Matplotlib, version 3.3.4~\citep{Hunter2007}.
Figure~\ref{fig:PDF_grid} shows the results for PV and wind capacity factors, and intra-day prices, respectively.

On the linear scale, the GAN and WGAN-generated scenarios do not show any considerable divergence from the historical PDFs. Meanwhile, clear differences can be observed for the wind capacity factor and the intra-day price scenarios generated by the VAE. 
Thus, the PDF identifies the shifted distribution of VAE-generated wind capacity factor scenarios and the compressed distribution of VAE-generated intra-day price scenarios.
Furthermore, the logarithmic scaling of the intra-day price PDF reveals that the GAN and WGAN fail to reproduce the heavy tails of the PDFs, i.e., they do not reproduce the rare price peaks that are present in the historical intra-day price time series. For both PV and wind capacity factors, the analysis of the PDFs in logarithmic scaling does not offer any additional insight into the distributions.  
Since most authors exclusively evaluate the linear scale, we emphasize the importance of analyzing the PDF on both linear and logarithmic scales, in particular, since the extreme events located in the heavy tails can critically affect system stability.
PV and wind capacity factor distributions have finite support and no heavy tails. Hence, it is sufficient to consider the linear scale.

The marginal PDFs of the generated data show worse matches than the full PDFs, even for the distributions without heavy tails. For instance, all DGMs learn poor fits of the bimodal structure of the marginal PV capacity factor PDF, i.e., the marginal distributions are not represented well although the full distribution indicates a good match.
The marginals of the VAE-generated intra-day price scenarios also display a much narrower distribution than the historical scenarios, which indicates that the VAE does not reproduce the full range of days with high and low prices. 
Note that this result is different from the also narrow intra-day price PDF, as it assesses the range of days with overall high or low prices as opposed to only the peaks.

\subsection{Autocorrelation function}\label{sec:autocorrelation}
\begin{figure*}[t]
    \includegraphics[width=\textwidth]{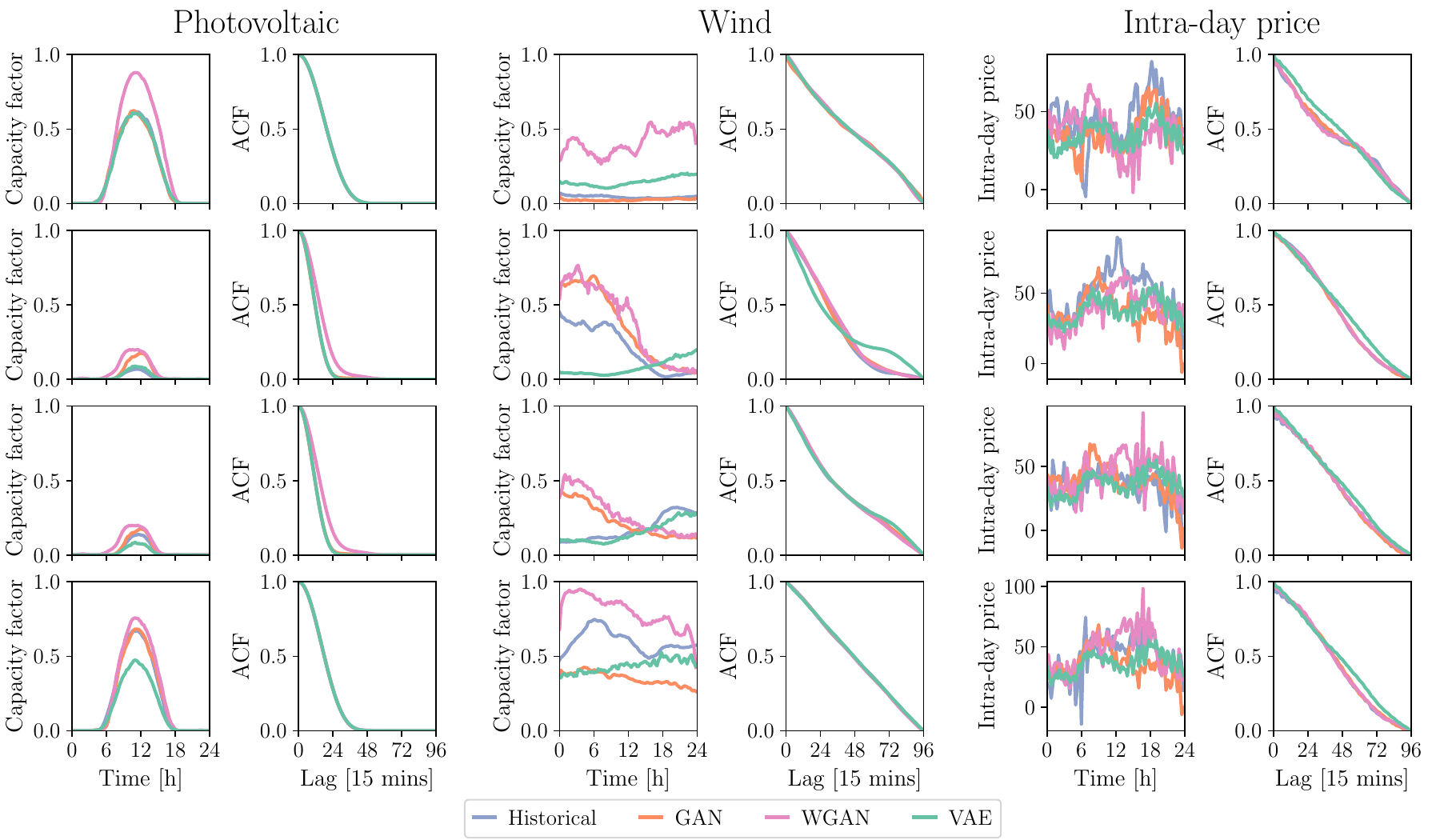}
    \caption{Autocorrelation function of selected historical scenarios and generated scenarios with Euclidean distance matching.
    Four exemplary scenarios of PV and wind capacity factor [-], and intra-day price [EUR/kWh] scenarios. Historical scenarios \citep{DataSource,DataSourcePrice} are compared to GAN-, WGAN-, and VAE-generated scenarios.}
    \label{fig:ACF}
\end{figure*}
For the analysis of the ACF, we follow the standard approach used in the literature of matching historical and generated scenarios by the best match of the ACF, see Section~\ref{sec:TheoryACF}.
Figure~\ref{fig:ACF} shows four selected scenarios and their respective ACFs for PV capacity factor, wind capacity factor, and intra-day prices, respectively. 

For all three data sets, the Euclidean distance selection described in Section~\ref{sec:TheoryACF} finds DGM-generated scenarios with an ACF similar to the ACF of the respective historical scenario, indicating a good match of the autocorrelation of the respective sample. 
However, as discussed in Section~\ref{sec:TheoryACF}, the matching ACFs are no proof of a good representation of the historical time series.
Indeed, Figure~\ref{fig:ACF} shows that there are scenarios in the generated sets that exhibit similar autocorrelation as the 4 considered historical scenarios. 
Besides, similar ACFs do not necessarily indicate that two scenarios stem from the same stochastic process, e.g., the first row of wind and the second row of price scenario in Figure~\ref{fig:ACF} have very similar ACFs despite describing two different data sets.
In conclusion, we find that the ACF should only be tested if a matching autocorrelation is a specific requirement. In particular, the limitation to single scenarios and the resulting lack of subsequent implications about the full scenario sets are highly problematic. 
In any case, the results should be treated with caution.

\subsection{Power spectral density}\label{sec:ExamplePSD}
\begin{figure*}[t]
    \centering
    \includegraphics[width=\textwidth]{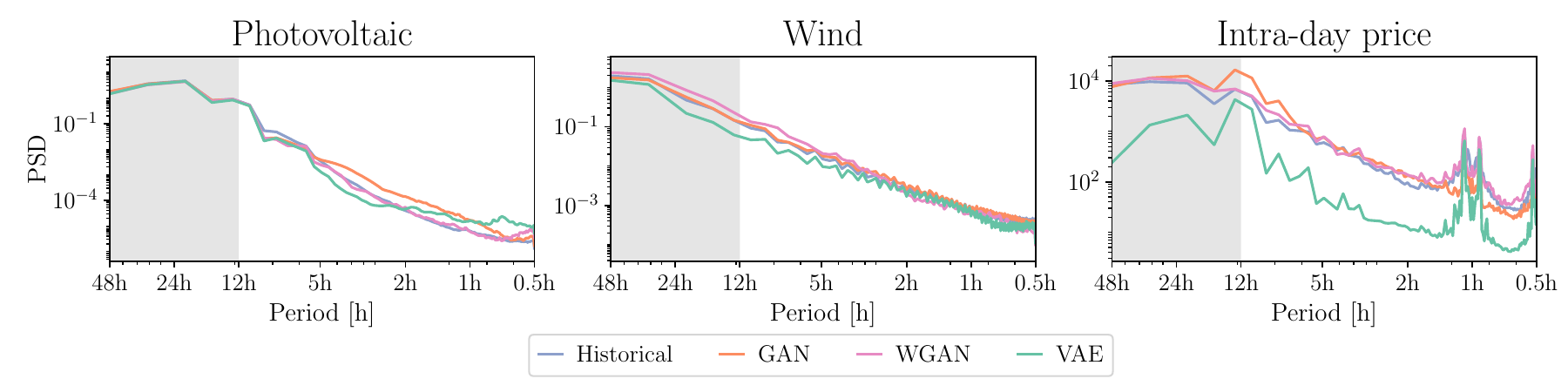}
    \caption{Power spectral density from Welch's method~\citep{welch1967use} of PV and wind capacity factor, and intra-day price scenarios. Historical datasets \citep{DataSource,DataSourcePrice} are compared to GAN-, WGAN-, and VAE-generated scenarios. Periods with potentially overlapping scenarios marked by gray areas.}
    \label{fig:PSD}
\end{figure*}
Following the standard literature approach, we concatenate the generated scenarios to form a single continuous time series and then calculate the PSD using Welch's method~\citep{chen2018model, wei2019short, qi2020optimal}. 
The results for PV capacity factor, wind capacity factor, and intra-day prices are shown in Figure~\ref{fig:PSD}.
We observe that the scenarios generated by the WGAN exhibit similar PSDs as the historical scenarios for all three data sets. 
The GAN- and VAE-generated scenarios show slight mismatches for PV and wind capacity factors, and there is a larger mismatch for the VAE-generated intra-day price scenarios. Specifically, the PSDs for the PV capacity factors reveal that the VAE-generated scenarios underestimate the fluctuations of medium length periods between 8\,h and 2\,h, and result in higher fluctuation for shorter time series, i.e., noisy scenarios. Meanwhile, the GAN-generated PV capacity factor scenarios have constantly higher fluctuations compared to the historical data. 
None of the above insights can be retrieved by relying on PDF and ACF alone. 

We present the PSD over periods between 48\,h and 30\,min, where 30\,min is the Nyquist frequency for the 15\,min sampling interval.
We mark the periods over 12\,h in gray to indicate that they are likely to overlap different scenarios and lead to spurious results.
Shifting the considered periods of our data proportional to \cite{chen2018model} as discussed in Section~\ref{sec:TheoryPSD} would result in periods between 3 weeks and 5\,h. 
Such a shift disregards, for instance, the peaks in the intra-day price PSD (see Figure~\ref{fig:PSD}).
Therefore, we conclude that it is indispensable to constrain the PSD to relevant periods, especially when working with concatenated time series.

\subsection{Multifractal Detrended Fluctuation Analysis}\label{sec:mfdfa}
\begin{figure*}[t]
    \includegraphics[width=1\textwidth]{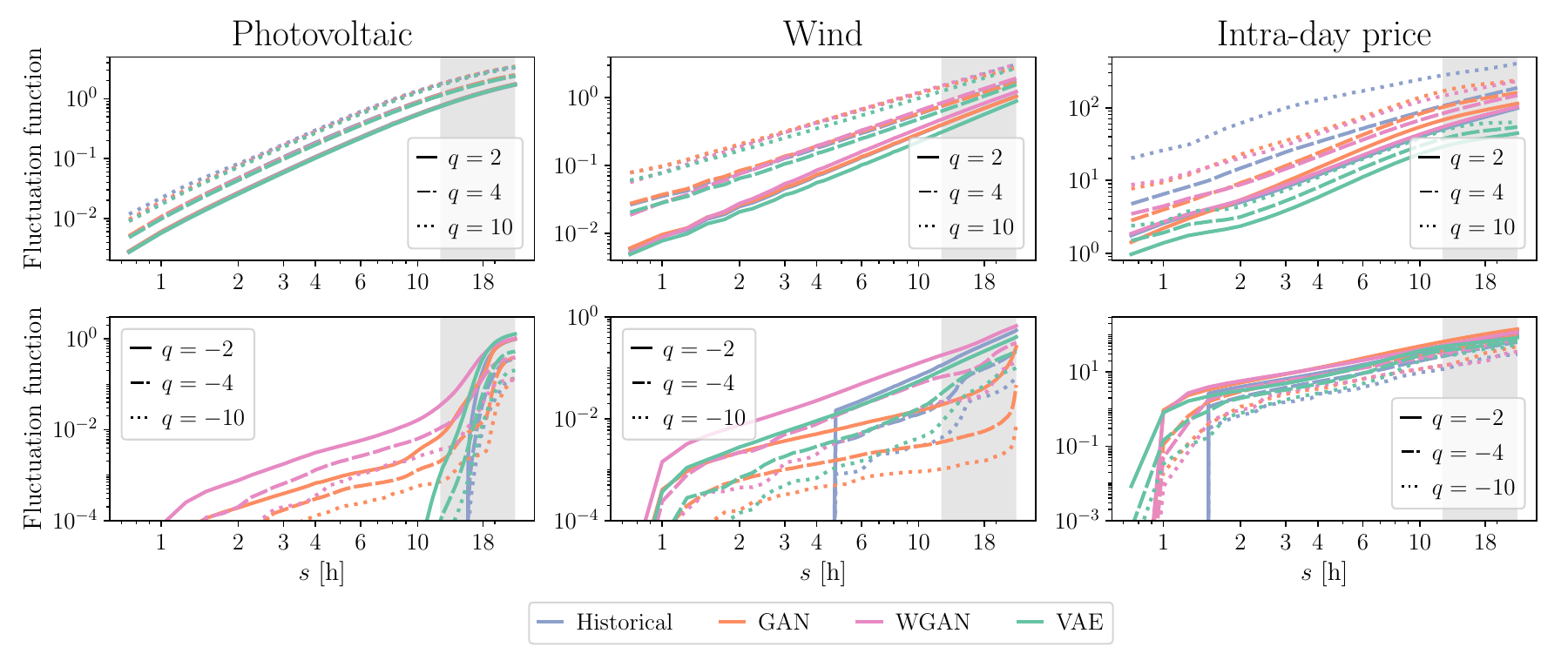}
    \caption{
    MFDFA of PV and wind capacity factor, and intra-day prices. Shown is the fluctuations function $F_q^2(s)$ with positive powers $q=2$, $q=4$, and $q=10$ (upper row) and negative powers $q=-2$, $q=-4$, and $q=-10$ (lower row). Historical data \citep{DataSource,DataSourcePrice} is compared to GAN, WGAN, and VAE-generated scenarios. Segment length with potentially overlapping scenarios marked by gray area.
    }
    \label{fig:MFDFA_grid}
\end{figure*}

We compute the fluctuation function over concatenated scenarios using linear detrending and $q$-powers $q=2$, $q=4$, $q=10$, and $q=-2$, $q=-4$, $q=-10$, respectively. The code for the MFDFA is available on GitHub~\citep{RydinMFDFA}.
Figure~\ref{fig:MFDFA_grid} shows the results for PV and wind capacity factor, and intra-day price scenarios.
We mark in gray segment lengths $s$ greater or equal to half the scenario length that are likely to overlap multiple scenarios and potentially result in spurious results.
For the positive $q$-powers, all the DGM-generated scenarios exhibit virtually no differences in the fluctuation function compared to the historical PV time series. 
For the wind capacity factor, the generated scenarios match the multifractal behavior of the historical time series, indicated by the change in slope for different $q$-powers. 
However, the fluctuation functions for WGAN with $q=2$ and $q=4$ show steep slopes that indicate stronger fluctuation in the long-term behavior, i.e., more rapid changes in the time series compared to the historical data. 
In the DGM scenario generation literature, the ACF is used to retrieve such deviations in the autocorrelation. 
However, in our application, the ACFs of the 4 analyzed scenarios do not indicate a false autocorrelation. 
Only MFDFA is able to achieve this task among the here considered validation methods.
For the intra-day price, the fluctuation functions of the DGM-generated scenarios deviate strongly from the historical time series, except for the $q=2$ fluctuation function of the GAN and WGAN scenarios. 
While the PSDs of GAN and WGAN indicate a good representation of the fluctuations in the intra-day price data, the MFDFA reveals that the DGMs capture only the average fluctuations, i.e., the PSD fails to retrieve the missing bursts and peaks in the generated scenarios. 

The MFDFA analysis with negative powers shows a steep drop-off by the fluctuation function for all three historical time series due to the data precision of two decimals for PV and wind capacity factor and three decimals for the intraday price. 
None of the DGM-generated scenarios match the drop-off and instead show fluctuation functions as a result of the untruncated values.
In the case of the PV scenarios, the drop-off can also be attributed to the constant period of zero production during night times. Here, the higher fluctuations indicate some noise in the data that was already shown by the PSD, see Section~\ref{sec:ExamplePSD}.
The analysis of wind scenarios with negative $q$-powers over longer segments highlights that WGAN-generated scenarios have high long-term, low amplitude noise around the plateaus.
Contrarily, GAN-generated scenarios display lower long-term noise than the historical data. 
We argue that peaks, bursts, and plateaus are essential elements of the process that need to be considered. Restricting the analysis to PDF, ACF, and PSD disregards these non-trivial elements and can lead to false conclusions about the generated scenarios.

  \section{Conclusion and Future Work}\label{sec:Conclusion}
\subsection{Summary}
We discussed commonly used validation methods for DGM-generated scenarios and their application in the literature with a special focus on energy time series. 
Our assessment focused on commonly used validation methods based on the probability density function (PDF), autocorrelation function (ACF), and power spectral density (PSD). 
In addition, we proposed using multifractal detrended fluctuation analysis (MFDFA) as a validation method for more complex features of time series like peaks, bursts, and plateaus.
Our assessment is specifically concerned with the validation of identically and independently distributed (iid) scenario sets and discusses the different validation methods in this particular context. 

Each of the four validation methods analyzes a particular feature of the generated data. 
For instance, the PDF describes the likelihood of the time series taking a certain value, while ACF, PSD, and MFDFA consider the relation of time steps in their order of occurrence. 
The literature approach of analyzing ACF only considers 3-4 single scenarios and thereby fails to give insight into the full DGM-generated scenario set. 
Furthermore, there is no guarantee that scenarios with matching ACF stem from the same stochastic process. 
Thus, the ACF results must be treated with caution.
In contrast, PDF, PSD, and MFDFA analyze the entire data set and thereby allow to derive conclusions about DGMs ability to generate realistic scenarios.

Between PDF, PSD, and MFDFA there is no one exclusive validation method to analyze the generated scenarios.
For instance, the PDF only considers the data distribution and neglects the correlation of time steps. 
The PSD retrieves insight into fluctuation and periodicity but considers neither the distribution nor the different scales of the stochastic process. 
The MFDFA neither assesses the distribution of the data nor the periodic behavior. However, MFDFA is uniquely able to dissect the different scales of the time series and retrieve, e.g., the missing bursts and peaks in the generated intra-day price scenarios. 
Analyzing the data using PDF, ACF, and PSD disregards these non-trivial elements, which can lead to false conclusions about the generated scenarios.
To summarise, no method precludes using other methods as well, and only the complement of multiple methods allows for an assessment of the generated data from different perspectives. 

\subsection{Practical implications}
In practice, requirements from the envisaged application of the scenarios should be used to select a set of appropriate validation methods.
For instance, the PDF ranks most relevant if the application requires the scenarios to represent the full range of possible realizations. 
If the scenarios have to include the extreme peaks of, e.g., the intra-day price, only MFDFA can give the relevant insight. 
Some validation methods can be omitted if the data is known to not include the analyzed features, e.g., it appears unnecessary to show the logarithmic scale PDF of the PV data considered in this work as the distribution does not have heavy tails. 
Regardless of the downstream application, we stress that all validation methods must be applied and interpreted correctly, e.g., considering the short time period characteristic of scenarios. 
For instance, PSD and MFDFA are based on concatenated scenarios and should not be evaluated for intervals longer than the scenario length.
Similarly, PDFs with heavy tails should be evaluated in logarithmic scaling in addition to the standard linear scale. 
In general, we do not expect the computational complexity of any of the validation methods to be prohibitive as they are typically applied only once after the training of the DGMs. In any case, PDF and PSD can be computed very efficiently using kernel density estimation and Welch's method, respectively, and the ACF only has to be computed for very few time steps. MFDFA is more complex compared to the other methods, however, it can be computed in reasonable times.

\subsection{Open research questions}
We stress that the list of validation approaches presented here is non-exhaustive, and other not included tests may offer valuable insight as well and should be selected depending on the requirements of the application. Furthermore, there are many open research questions in the field of scenario validation. 

Importantly, PDF, ACF, PSD, and MFDFA are all based on visual inspection of their respective outputs, which makes their interpretation subjective and potentially error-prone. 
This is further highlighted by our discussion and we argue that naive interpretation of the validation results can be susceptible to spurious conclusions. 
Consequently, future research on scenario validation should focus on deriving quantitative metrics that allow for an easier and comparable interpretation. A representative set of such quantitative metrics could then result in a comprehensive taxonomy allowing for consistent and reproducible validation of generated scenario sets. 
Furthermore, quantitative metrics could be used to better design the DGMs, e.g., in hyperparameter tuning. 

As emphasized throughout our paper, this work concerns only the validation of iid DGM-generated scenarios. However, scenarios obtained with other machine-learning models may require different validation approaches. For instance, autoregressive models \citep{vagropoulos2016ann} generate scenarios based on inputs, which makes each set of scenarios specific to those inputs and, thereby, nontrivial to compare.

Another important aspect of validation is multivariate analysis, i.e., the validation of scenarios that include multiple potentially correlated stochastic processes at the same time. The four validation methods discussed in this work are all inherently univariate, which makes their application to multivariate time series impossible. Therefore, multivariate analysis ranks highly important as there are strong correlations between different energy time series, e.g., renewable power generation and electricity prices \citep{weron2014electricity}.

Besides improved analysis of the data itself, the application of the generated scenarios must be considered. For instance, stochastic programming \citep{morales2013integrating} is a highly relevant problem in many energy system circumstances. Here, a set of scenarios is used to discretize the true probability distribution of the stochastic processes, and different sets of selected scenarios should lead to consistent objective function values \citep{kaut2003evaluation}.

\subsection{Conclusion}
In conclusion, validation methods for DGM-based scenario generation should receive more attention.
In general, we find that relying on single validation methods is not sufficient and the generated scenarios should always be analyzed using multiple methods. 
Research on DGM-based scenario generation should emphasize accurate application and interpretation of methods that detail the relevant properties in the data they wish to reproduce.
With the lack of loss functions that detail the quality of the generated data, scenario validation can be a valuable asset in the design and testing of DGM model structures.
Furthermore, future development of new validation methods should focus on the application of scenarios in downstream decision-making, e.g., in numerical optimization.

  \section*{Acknowledgements}
\label{sec:acknowledgements}
This work was performed as part of the Helmholtz School for Data Science in Life, Earth and Energy (HDS-LEE) and received funding from the Helmholtz Association of German Research Centres, amongst others, via the grant \textit{Uncertainty Quantification -- From Data to Reliable Knowledge (UQ)}, with grant no.~ZT-I-0029.
This project has received funding from the European Union’s Horizon 2020 research and innovation program under the Marie Sklodowska-Curie grant agreement No 840825.

  \bibliographystyle{apalike}
  \renewcommand{\refname}{Bibliography}
  \bibliography{literature.bib}

\end{document}